\documentclass[pdflatex,sn-mathphys-num]{sn-jnl}

\usepackage{graphicx}%
\usepackage{multirow}%
\usepackage{amsmath,amssymb,amsfonts}%
\usepackage{amsthm}%
\usepackage{mathrsfs}%
\usepackage[title]{appendix}%
\usepackage{xcolor}%
\usepackage{manyfoot}%
\usepackage{booktabs}%
\usepackage{algorithm}%
\usepackage{listings}%
\usepackage{algorithmic}

\theoremstyle{thmstyleone}%
\theoremstyle{thmstyletwo}%

\theoremstyle{thmstylethree}%

\begin{document}

\title[Article Title]{\textit{Ex vivo} breach detection using electrical conductivity during robotic pedicle drilling in the spine}










%
\author[1,2]{\fnm{Jorge Andres} \sur{Perez Velasquez}}

\author[2]{\fnm{Fran\c cois} \sur{Teyssere}}

\author[2]{\fnm{Thibault} \sur{Chandanson}}

\author[3]{\fnm{Quentin} \sur{Grimal}}

\author[1]{\fnm{Brahim} \sur{Tamadazte}}\email{brahim.tamadazte@cnrs.fr}

\affil*[1]{\orgdiv{ISIR}, \orgname{Sorbonne Universit\'e, CNRS UMR 7222, Inserm U1150}, \orgaddress{\street{Place de Jussieu}, \city{Paris}, \postcode{75005}, \state{} \country{France}}}

\affil[2]{\orgdiv{SpineGuard SA}, \orgname{}, \orgaddress{\street{10 cours Louis Lumi\'ere}, \city{Vincennes}, \postcode{94300}, \state{}  \country{France}}}

\affil[3]{\orgdiv{LIB}, \orgname{Sorbonne Universit\'e, CNRS UMR 7371, Inserm U1146}, \orgaddress{\street{15 rue de l'\'Ecole de M\'edecine,}, \city{Paris}, \postcode{75006}, \state{} \country{France}}}


\abstract{\textbf{Purpose:} Pedicle screw placement is technically demanding in scoliosis treatment. High precision is required due to limited visibility, anatomical variability, and the risk of complications. Although robotic systems assist CT-based planning and execution, they still rely on ionizing intraoperative imaging and complex registration. This study proposes robotic pedicle drilling with real-time preventive breach detection using electrical bioimpedance sensing.\\

\textbf{Methods:} We developed a robotic approach combined with a pedicle-drilling tool equipped with a proprioceptive electrical bioimpedance sensor developed by SpineGuard\textsuperscript{\textregistered}. A real-time detection algorithm was designed to analyze the electrical bioimpedance signal during drilling and identify abrupt changes in conductivity associated with potential breaches towards the spinal canal. The method operates without external devices or sensors.\\

\textbf{Results:} The ex vivo experiments showed that the proposed method prevented breaches in 100\% of the 51 drilling cases. These findings demonstrate the system's ability to detect potentially hazardous events during drilling and to stop the procedure before. The ex vivo experiments demonstrated that the proposed method prevented breaches in all 51 drilling cases.\\

\textbf{Conclusions:} This work demonstrates the feasibility of robotic pedicle drilling with electrical bioimpedance sensing for real-time breach prevention. Using only the tool signal, the method eliminates the need for external sensing systems and supports safer pedicle screw placement.}
 



\maketitle

\section{Introduction}\label{sec.introduction}
\begin{figure}
\centering
\includegraphics[width=.6\columnwidth]{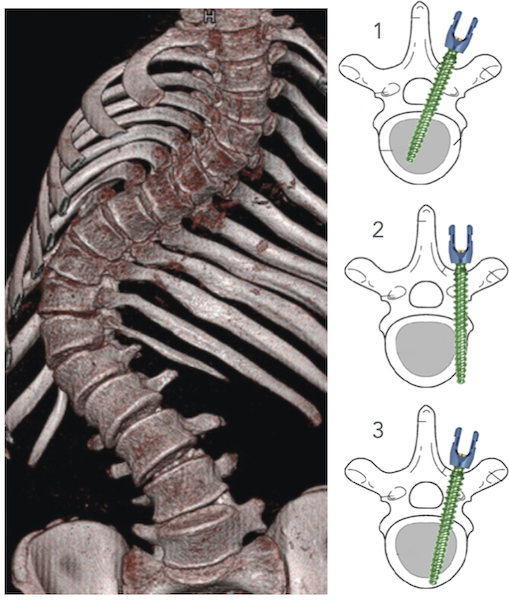}
\caption{Left: a typical scoliose deformation. Right: different scenarios for screw insertion into the vertebrae: (1) leading to medial breach, (2) leading to lateral breach, and (3) well-positioned pedicle screw.}
\label{fig.scoliose}
\end{figure}

%
%
Pedicle Screw Placement (PSP) is a surgical procedure used to treat various spinal pathologies, such as scoliosis, spinal trauma, or spinal instability. Drilling the pedicles for screw positioning is a challenging procedure. Failure can lead to injuries, such as major vessels (aorta, vena cava) or the spinal cord~\cite{esses1993complications}. The literature reports a complication rate in the case of screw mispositioning of up to 18\%~\cite{tang2014position}. Indeed, the main risk arises during the pedicle preparation phase, when the surgeon defines a drilling trajectory that is considered optimal and safe for the patient. The surgeon has no means of real-time, in-depth visualization of the vertebra and therefore relies primarily on skill and experience. For example, in scoliosis (Fig.~\ref{fig.scoliose} left), which is characterized by severe three-dimensional deformities of the spinal column, identifying the entry point for each pedicle screw is particularly challenging. This often leads to high rates of pedicle screw misplacement, and approximately 7\% of these mispositioned screws may result in medial or lateral breaches (Fig.~\ref{fig.scoliose}, right). Lateral breaches can damage vascular structures, causing significant bleeding and potentially life-threatening complications~\cite{lopera2010aortoiliac}. When medial breaches occur into the spinal canal, they can expose the spinal cord to risk and lead to neurological consequences ranging from paraparesis to complete paralysis~\cite{mac2013neurological}. The literature reports that most breaches are classified as \textit{"A"} or \textit{"B"} under the Gertzbein and Robbins classification~\cite{gertzbein1990accuracy}. In this well-established classification, the position of a transpedicular screw is graded from \textit{"A"} to \textit{"E"} based on the extent of pedicle cortex breach: grade \textit{"A"} corresponds to a fully intrapedicular position without cortical violation, grade \textit{"B"} to a breach of less than 2~mm, grade \textit{"C"} to a breach of 2–4~mm, grade \textit{"D"} to a breach of 4–6~mm, and grade \textit{"E"} to a breach greater than 6~mm or a screw located outside the pedicle.

To improve the accuracy of pedicle screw placement and prevent breaches, various techniques have been proposed in the literature. First, patient-specific 3D-printed guides generated from preoperative imaging have been introduced as templates to assist PSP procedures~\cite{sheha20193d}. Several studies also report the use of different non-visual exteroceptive or proprioceptive sensors, including vibroacoustic, accelerometric, torque, and force sensors~\cite{seibold2021real, massalimova2023automatic, saghbiny2024breach}. However, most solutions rely on image-based navigation, using fluoroscopy or CT scans. 

Furthermore, robotic assistance in spinal surgery was first introduced in 2004, marking a major advancement in the field, which primarily uses navigation systems to optimize pedicle screw placement~\cite{bozic2003economic}. Most of the available robotic systems are integrated with stereotactic navigation systems~\cite{kia2021robotic}. The SpineAssist system, developed by Mazor Robotics, was the first robotic system for spinal surgery to receive FDA (Food and Drug Administration) approval. Over the years, significant advances have been achieved, with the latest generation, Mazor~X\textsuperscript{\textregistered}, incorporating a camera to assess the surgical environment and improve accuracy~\cite{ringel2012accuracy}. Another widely used robotic system is ROSA Spine\textsuperscript{\textregistered}, which operates according to similar shared-control principles~\cite{lefranc2016evaluation}. The most recent system, ExcelsiusGPS\textsuperscript{\textregistered}, integrates real-time intraoperative imaging and enables screw placement without the use of guidewires~\cite{zygourakis2018technique}.

Although navigation and robotic systems have improved spinal surgery, they remain limited by usability, cost, and accuracy. Respiratory motion and small patient movements during pedicle preparation can introduce substantial registration errors~\cite{shuman2022intraoperative}. Moreover, these systems improve trajectory planning but do not fully prevent pedicle breaches. To address this limitation, we developed a robotic platform for pedicle screw placement that integrates a SpineGuard\textsuperscript{\textregistered} bioelectrical conductivity sensor, hereafter referred to as the DSG sensor, which is already used clinically~\cite{bai2013comparison, yurube2022improved}. A threaded drill bit with a pyramidal sensorised tip operating at 25 Hz was mounted on a custom motorized drilling system attached to Kurobot’s end effector.

A real-time algorithm uses adaptive conductivity thresholds and a dual-alert mechanism to stop drilling before a breach occurs. The system was evaluated ex vivo on fresh pig vertebrae through 51 drillings, achieving preventive detection in all trials. The mean error remained within [-1, +1.4]~mm, and 100\% of drillings were graded \textit{"A"} or \textit{"B"} according to the Gertzbein and Robbins classification.

\section{Materials and Methods} \label{sec.method}
\subsection{Experimental Setup}
%
%

\begin{figure}[!h]
\centering
\includegraphics[width=0.95\columnwidth]{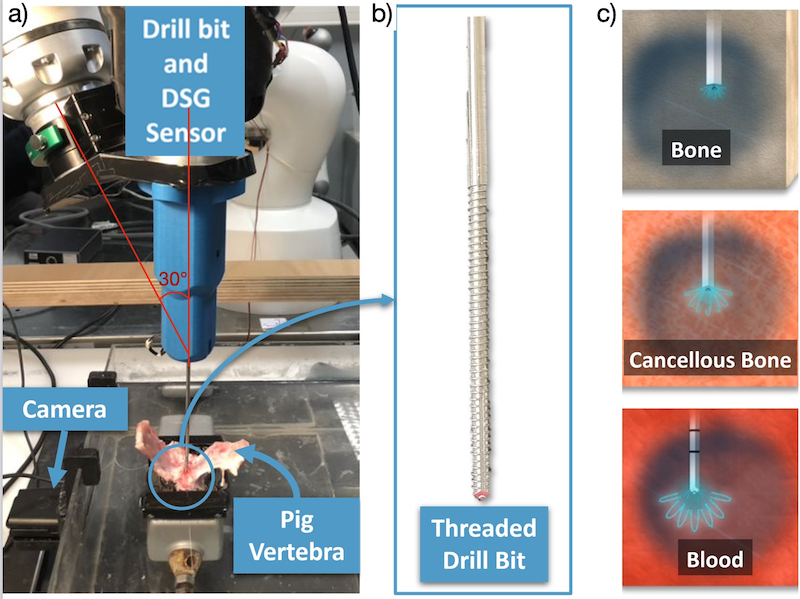}
\caption{From left to right: detail of the drill placed at the entry point, detail of the threaded drill bit, behavior of electrical conductivity in different tissues}
\label{fig:DSG}
\end{figure}

The SpineGuard\textsuperscript{\textregistered}-derived drill bit (Fig.~\ref{fig:DSG}(a)) has a 3~mm diameter and a pyramidal tip with bipolar electrodes that provide a real-time electrical signal dependent on the tissue traversed. Its threaded profile maintains contact with the bone and limits slipping along the drilling axis (Fig.~\ref{fig:DSG}(b)). The drill is driven by a custom unit delivering 1.5~Nm nominal torque and up to 922~rpm, and is mounted on the robot end-effector at 30$^\circ$ to improve surgical access. DSG technology measures local tissue conductivity using the bipolar electrodes~\cite{0}: conductivity is low in cortical bone, higher in cancellous bone, and highest in blood and soft tissue (Fig.~\ref{fig:DSG}(c)).

%

\begin{figure}
    \centering
    \includegraphics[width = 0.9\columnwidth]{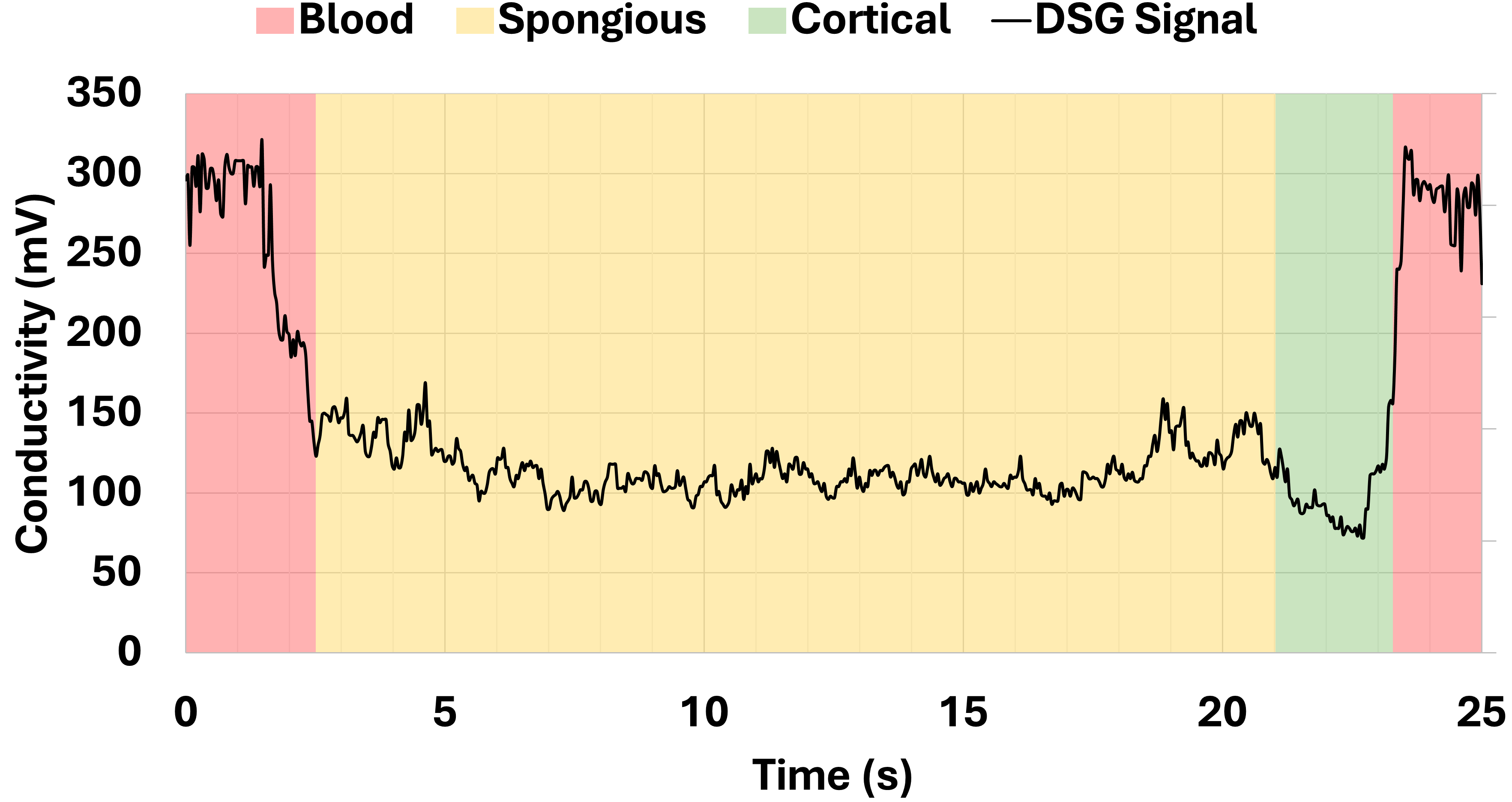}
    \caption{A typical DSG signal acquired during a vertebra drilling process.}
    \label{fig.DSG_Signal}
\end{figure}

Figure~\ref{fig.DSG_Signal} shows a representative DSG signal acquired during drilling in a fresh pig vertebra. Tissue-dependent conductivity variations enable detection of both lateral and medial breaches. Micro-fractures in the cortical bone allow fluid to reach the pyramidal tip, producing a detectable signal change before the drill fully enters the spinal canal.

%
\subsection{Drilling Workflow}
To mimic the conditions of a typical drilling procedure on a patient, fresh pig vertebrae were immersed in a saline solution that reproduces the electrical conductivity of cerebrospinal fluid (CSF). This solution contains sodium chloride (NaCl), chosen to replicate the ionic composition of CSF, particularly the sodium (Na\(^+\)) and chloride (Cl\(^-\)) ions, which play a crucial role in its conductivity. The concentrations of these ions in CSF are approximately 130--150 mEq/L for sodium and 120--125 mEq/L for chloride~\cite{10}. To replicate this ionic environment, which reflects a near 1:1 ratio of sodium and chloride ions, we prepared a solution with a NaCl concentration of approximately 120--130 mmol/L. 

%
\subsection{Vertebrae Collection and Preparation}
Pig vertebrae exhibit a pedicle morphology similar to that of humans, making them a suitable model. A limitation is the presence of growth cartilage at the vertebral ends, which may affect conductivity signals. To minimize these artifacts, we selected a drilling trajectory from the pedicle towards the spinal canal, simulating a misjudged surgical path while avoiding the cartilage.


Compared with spinous process drilling~\cite{8}, pedicle trajectories may approach the spinal canal tangentially, reducing electrode contact with cerebrospinal fluid. Conductivity changes are therefore weaker and more gradual, requiring robust signal processing to reliably detect early breaches.

In this study, 51 fresh porcine lumbar vertebrae were prepared by a surgeon to simulate operating room conditions. Each vertebra was clamped in a transparent box facing a camera for continuous inspection of the spinal canal. An orthopedic surgeon created the entry point with an awl, after which the robot was hand-guided into position. Drilling was performed at 10~N and 30 rpm, with a maximum descent rate of 50 mm/s. The real-time breach-detection method is summarised in \emph{Algorithm 1}.
%
\subsection{Methodology}
To verify result accuracy and detect potential breaches, several approaches were applied both during and following the drilling experiments: 
\begin{enumerate}
    \item \textit{Palpation with a ball-tip feeler:} this method (usually used by surgeons during surgery) was used to assess the integrity of the cortical bone and rule out the existence of grade \textit{"C"} breaches.
    \item \textit{Visual inspection:} following the experiments, the spinal canal was visually examined to identify any deformation caused by the drill bit tip.
    \item \textit{Imaging and video analysis:} a combination of CT scans and video recordings obtained during drilling was used to confirm the absence of false positives and false negatives and to validate the experimental results.
\end{enumerate}

These methods ensured a comprehensive analysis of both the mechanical and structural impacts of the drilling process.
\subsection{Breach Detection Algorithm and Dataset}
\begin{figure}[!h]
\centering
\includegraphics[width =  \columnwidth]{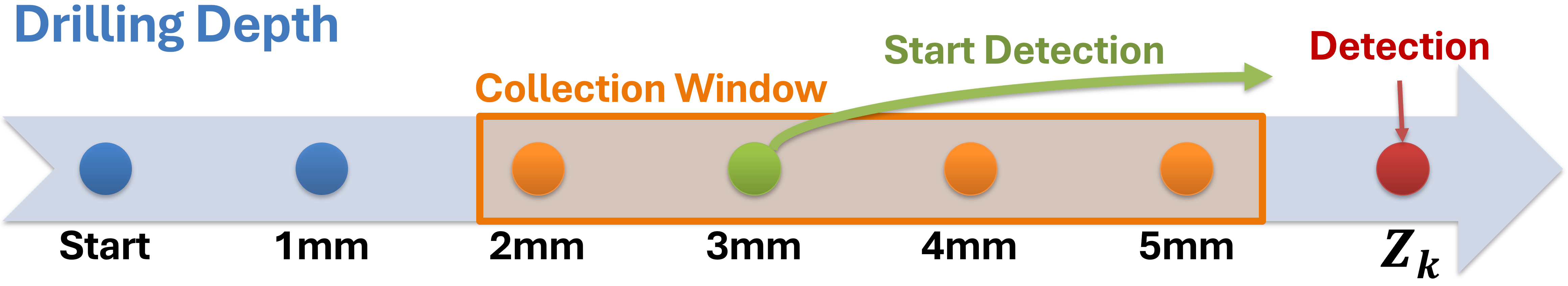}
\caption{Illustration of the drilling depth during data collection. The orange window (2–5~mm) denotes the measurement range, the green arrow indicates the detection threshold at 3 mm, and the red marker indicates a detection event.}
\label{fig:Graph_Depth}
\end{figure}

%
%
A dataset from prior pedicular drillings in pig vertebrae was used to tune the algorithm offline. It contained synchronized measurements of drilling depth and electrical conductivity, stored in time-stamped \textit{rosbag} files. Replaying and analyzing these data revealed how conductivity varied with drilling depth. Specifically, initial evaluations revealed that in thinner pedicles, breaches could occur before the end of the measurement window initially defined at $z_{end\_collect}=5~\text{mm}$~\cite{4}. These findings were instrumental in calibrating the program: the measurement window was adjusted to begin at $z_{collect}=2~\text{mm}$, and the alert was allowed to be triggered as early as $z_{detect}=3~\text{mm}$ depth (Fig.~\ref{fig:Graph_Depth}).

The proposed method operates as follows. First, an adaptive threshold, denoted ($\psi_r$), is computed from the conductivity measurements recorded over the first few millimeters of drilling. This initial segment is used as a baseline because it corresponds to engagement in predominantly cortical and cancellous bone, before any interaction with the spinal canal is expected.

To make the threshold robust across specimens and drilling conditions, ($\psi_r$) is constrained within predefined bounds. An upper bound ($\psi_{\max}$ = 500 $\mathrm{mV}$) is applied to avoid false negatives in cases where the initial conductivity values are already high, which would otherwise push the adaptive threshold upward and reduce sensitivity to subsequent changes. Conversely, a lower bound ($\psi_{\min}$ = 150 $\mathrm{mV}$) is used to limit false positives. In the dataset described above, the high density of porcine pedicles produced very low conductivity values, resulting in a low ($\psi_r$) and making the algorithm overly sensitive to normal signal fluctuations and measurement noise.

The algorithm uses conductivity measurements ($\psi$), stored in $\Psi$, and drilling depth ($z$), stored in $Z$. These paired data provide a depth-indexed trace of the electrical response at the drill tip. Using $\psi_r$ as a reference, the method detects signal variations associated with transitions from bone to fluid-filled regions and triggers a breach flag when the decision criteria are met.

During the initial data acquisition phase (the beginning of drilling), if $Z_k$ exceeds the predefined depth threshold $Z_{collect}$, the mean conductivity $\psi_r$ is updated as follows:
\begin{equation}
    \psi_r = \frac{1}{k-j} \sum_{i=j}^{k} \psi_i
\label{eq.psi_r}
\end{equation}
where $\psi_i$ is the conductivity in millivolts (mV) measured at instant $i$, the index $k$ corresponds to their last position in the acquired signal, and $j$ is the index of the beginning of the data collection for adaptive threshold calculation.

Next, during the breach detection phase, the threshold $\psi_t$ is computed as follows:
 \begin{equation}
   \psi_t = \max \Big(\min(\psi_{max}, \alpha \cdot \psi_r), \psi_{min}\Big)
\end{equation}
where $\alpha$ is a predefined coefficient that helps to weigh the threshold. The coefficient $\alpha = 2.4$ was empirically determined in an earlier study~\cite{4}.

This is followed by a comparison between the current conductivity $\psi_k$ and the computed threshold conductivity $\psi_t$. If  $\psi_k > \psi_t$, flag $A_1$ is set to $True$:
\begin{equation}
   A_1 \leftarrow \psi_k > \psi_t
\end{equation}

Next, the depth index $m$ closest to $\Delta z=2mm$ before the last recorded depth is estimated as follows:
\begin{equation}
\nonumber
    m \leftarrow \underset{i \in \{1, \ldots, k\}}{\text{argmin}} \left| Z(i) - (Z_k - \Delta z) \right|
\end{equation}
where $i$ is the index skimming through the recorded depth, $Z(i)$ is the depth at index $i$, and $\Delta z$ is the depth window within which we look for a conductivity gradient.

The minimum conductivity within the defined search window is then computed:
\begin{equation} 
\nonumber
   n \leftarrow \underset{i \in \{m, \ldots, k\}}{\text{argmin}} \Psi(i)
\end{equation}
where $\Psi(i)$ is the electrical conductivity  at index $i$.

The detection of a sudden change in the conductivity value, indicating the onset of a breach, is performed as follows:
\begin{equation}
    \underset{i \in \{n, \ldots, k\}}{\max} \Big( \Psi(i) - \Psi(n)\Big) > \Delta max
\end{equation}
where $\Delta max$ is the conductivity gradient threshold calibrated as $\Delta max=230mV$: if a variation of conductivity superior to this threshold is detected within the search window, flag $A_2$ is set to $True$. 

Finally, an alert is triggered if either $A_1$ or $A_2$ was set to $True$: 
\begin{equation}
    Alert \leftarrow A_1 || A_2 
\end{equation}

\begin{algorithm}[!h]
\caption{Bone breach detection.}\label{alg:alg1}
The algorithm is called for each new measurement; Initially \emph{Alert} = false, $\psi_r$ = 0, and $\Psi$, \emph{Z} are empty lists. \textcolor{blue}{Blue values} are the new or modified parameters.\\

\begin{algorithmic}
\STATE $\Psi$ $\leftarrow$ [$\Psi$ \quad $\psi_k$ ], \emph{Z} $\leftarrow$ [ \emph{Z \quad $Z_k$} ] 

\STATE \textbf{if} \emph{$Z_k$} $>$ \textcolor{blue}{$z_{collect}$} \textbf{then}\\

\quad \textbf{if} \emph{$Z_k$} $<$ \textcolor{blue}{$z_{end\_collect}$} \textbf{then} 
\\\vspace{0.1cm}
\quad \quad $\psi_r = \frac{1}{k-j} \sum_{i=j}^{k} \psi_i$

\textbf{else} \\
    \quad  \emph{j}$  \leftarrow$ \emph{k} \\

\textbf{end if}\\

\textbf{if} \emph{$Z_k$} $>$ \textcolor{blue}{$z_{start\_detect}$} \textbf{then}

\quad \quad \emph{$\psi_t = $} max(min({$\psi_{max}$}, $\alpha \cdot \psi_r$ ), \textcolor{blue}{$\psi_{min}$}) 
 
\quad \quad \emph{$A_1$} $\leftarrow$ $\psi_k$ $>$ $\psi_t$ 

$\quad \quad \emph{m} \leftarrow \underset{i\in \{ 1,\ldots,k\}}{\text{argmin}} \left| Z(i)-(z_k-\Delta z) \right|$\\

$\quad \quad \emph{n} \leftarrow \underset{i\in \{ m,\ldots,k\}}{\text{argmin}} \Psi(i) $ 

\quad \quad \emph{$A_2$} $\leftarrow ((\underset{i\in \{ n,\ldots,k\}}{\text{max}} \Psi(i))-\Psi(n)) >$ {$\Delta max$} \\

\quad \quad \emph{Alert} $\leftarrow$ \emph{$A_1$} $||$ \emph{$A_2$} \\

\textbf{end if}\\

\end{algorithmic}
\label{alg1}
\end{algorithm}

%
%
\section{Results}\label{sec.results}
%
The surgeon positioned the drill bit at the pedicle entry point by manually manipulating the robotic arm; thereafter, drilling was performed automatically. The applied contact force and the drill’s rotational speed were adjusted according to the vertebral bone quality. In general, a nominal drilling force of $f_d$ = 10 $\mathrm{N}$ was used; however, depending on the mechanical properties of the bone, $f_d$ can be increased up to 100 $\mathrm{N}$.

In total, 51 fresh lumbar pig vertebrae were drilled under conditions intended to approximate an operating room setting. Drilling was stopped online by the perforation-detection algorithm without any additional calibration. After each trial, the surgeon used a ball-tipped probe to assess the integrity of the pedicle walls and confirm the depth. Post hoc inspection revealed that, for all 51 vertebrae, drilling was interrupted immediately before perforation, resulting in a 100\% success rate, according to grade \emph{"A"} (i.e., fully intrapedicular position of the screw) and \emph{"B"} (i.e., the breach does not exceed 2~mm) according to the \textit{Gertzbein-Robbins} classification~\cite{gertzbein1990accuracy}. Grade \emph{"A"}  and grade \emph{"B"} screws are deemed clinically acceptable when evaluating screw placement post-operatively.

\begin{figure}
\centering
\includegraphics[width=0.99\columnwidth]{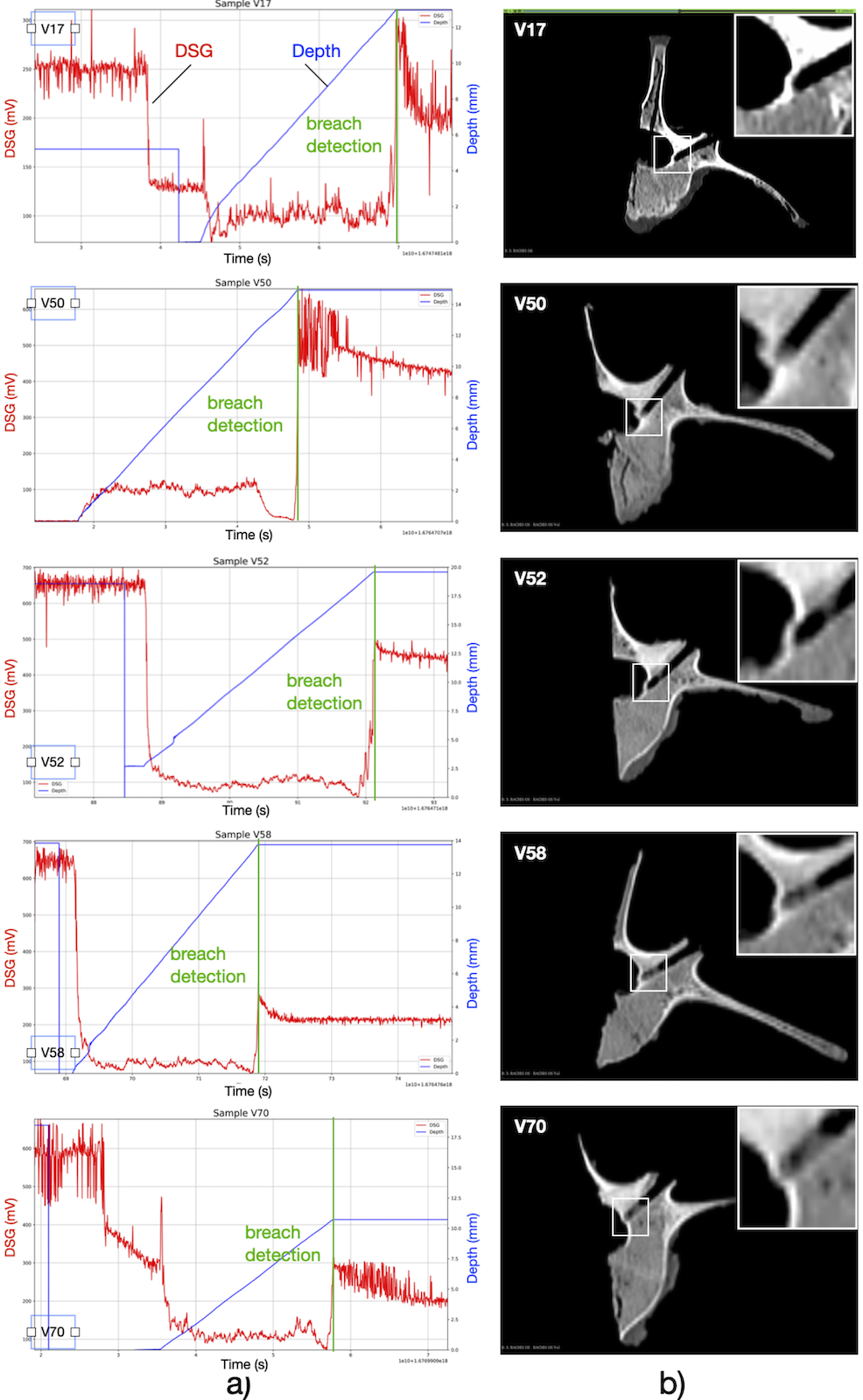}
\caption{Examples of breach-detection assessment. (a) Electrical conductivity, drilling depth, and detection time. (b) CT or $\mu$CT scans showing drill-tip deformation, with enlarged views of the onset of canal breach.}
\label{fig:Curve}
\end{figure}

Figure~\ref{fig:Curve}(a) shows the electrical conductivity traces for the drillings presented in Fig.~\ref{fig:Curve}(b). For instance, in vertebra V58, high conductivity is initially observed, likely due to contact with blood introduced during entry-point preparation. At this stage, the drill tip is not yet fully engaged with the bone, so the drilling process has not yet begun ($z_k < z_{detect}$). Once the tip enters cancellous bone, the conductivity decreases and then stabilizes at values characteristic of this tissue. Another example of electrical conductivity is observed in vertebra V50: the drill penetrates into cortical bone, which is less conductive, resulting in a decrease in the measured signal. Across all cases, a sharp conductivity peak is observed, corresponding to the onset of a micro-breach of the cortical layer, after which drilling is stopped. These peak events are marked by arrows in Fig.~\ref{fig:Curve}(a).

\begin{figure}
\centering
\includegraphics[width=.7\columnwidth]{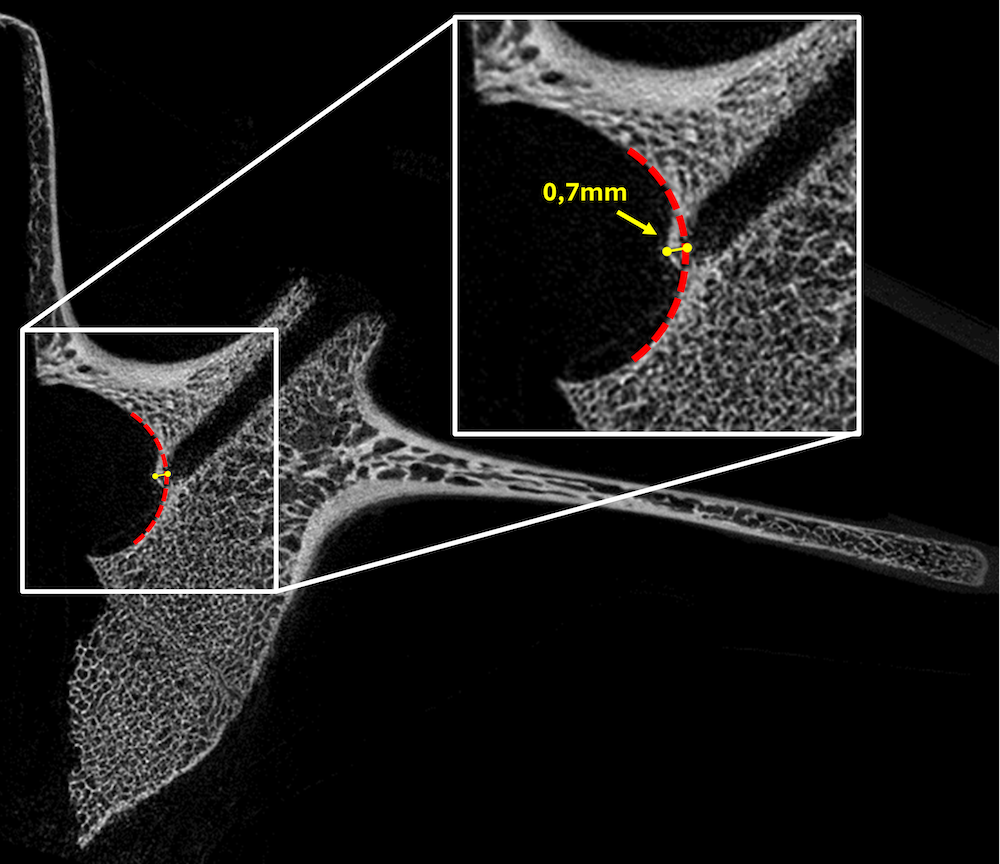}
\caption{Micro-CT scan showing the spinal canal boundary (red dotted line) used to measure cortical deformation, indicated by the yellow line.}
\label{fig.erreur_breche}
\end{figure}

\begin{figure}
 \centering
 \includegraphics[width=\columnwidth]{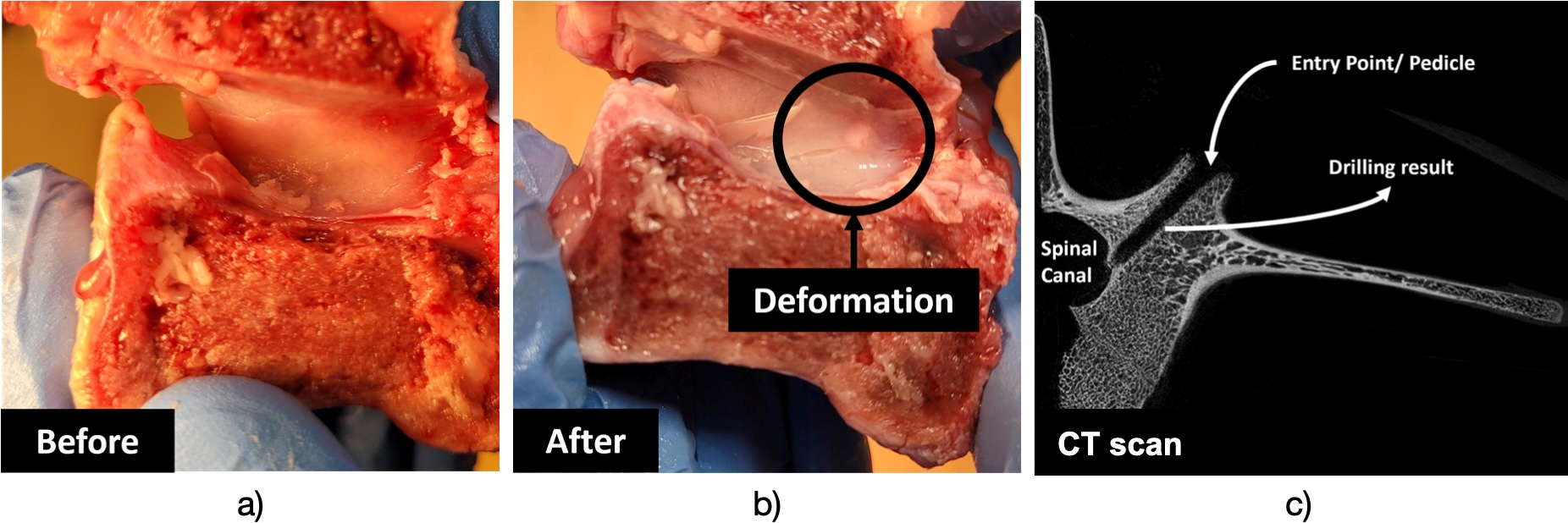}
 \caption{RGB images and micro-CT scans of the spinal canal before drilling (a) and after drilling (b). A slight canal deformation is visible when the method automatically stops drilling. Panel (c) shows this non-perforating deformation on the micro-CT scan. }
 \label{fig.perforation}
 \end{figure}

\begin{figure}
\centering
\includegraphics[width=0.9\columnwidth]{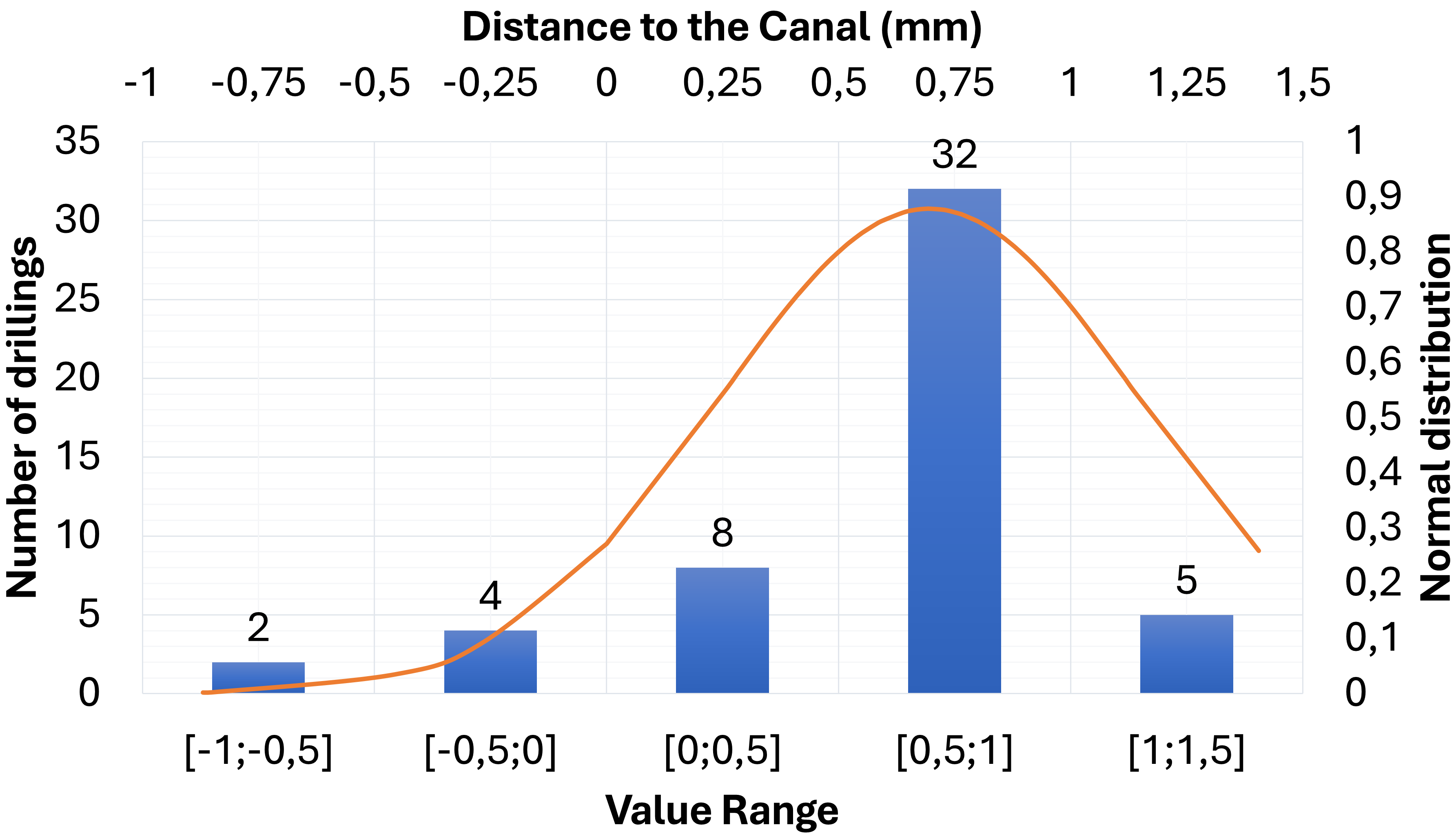}
\caption{Histogram showing the deformation caused by the drill tip at various distances from the canal. The peak indicates the most frequent deformation range [0.5–1 mm].}
\label{fig:Statistics}
\end{figure}

The deformation was measured using 3D Slicer, an open-source software package for CT scan visualization (Fig.~\ref{fig:Curve}(b)). To measure the deformation, the outer layer of the spinal canal was taken as the reference point. Then, a line was projected in the direction of the spinal canal, yielding a positive deformation length (Fig.~\ref{fig.erreur_breche}). If there is no deformation, the drill stops before deforming the spinal canal, resulting in a negative deformation length. The mean length measured was 0.7~mm, with a standard deviation of 0.45~mm. All measured lengths were in a range of –0.9~mm to +1.4~mm (Fig.~\ref{fig:Statistics}). As illustrated in this figure, the distribution of stopping errors (i.e., the deformation length of the bony canal) is as follows: 6 drillings are classified as grade \textit{“A”} and 45 as grade \textit{“B”}, with none coming close to the 2~mm threshold that would result in a downgrade to grade \textit{"C"}. These findings show that the breach detection algorithm, when used with the electrical conductivity sensor, is both precise and reliable, consistently halting drilling early enough to prevent perforation of the canal, as shown in Fig.~\ref{fig.perforation}. 

%
\section{Discussions}\label{sec.discussion}
%
An initial validation of this work was first carried out on lamb vertebrae, whose size is comparable to that of human vertebrae~\cite{4}. The robotic platform and the breach detection algorithm were evaluated under favorable conditions by performing drillings along the spinous process (a perfectly vertical drilling trajectory), rather than within the pedicles. This setup enabled us to verify the robot’s force controller, the speed control of the drilling unit, the relevance of using a threaded drill bit instead of a smooth one, and the overall quality of the electrical conductivity signal. An initial version of the breach detection algorithm was also evaluated in these experiments.

The results showed high detection accuracy and success in breach identification~\cite{4}, motivating improvements to the robotic platform and a more robust breach-detection algorithm. We then tested the system under more clinically representative conditions by drilling the left and right pedicles of fresh pig vertebrae, whose morphology is similar to that of humans~\cite{6,9}. To isolate breach detection, only drillings following the planned trajectory towards the spinal canal were analyzed; clinically unacceptable deviations were excluded.

Future studies should consider growth cartilage in porcine specimens, particularly for pedicle-to-vertebral body trajectories. Its higher conductivity may confound breach detection and cause false positives. This can be addressed by characterizing its conductivity and incorporating it into the decision logic, or by using mature specimens without growth cartilage.

\section{Conclusions}\label{sec.conclusion}
%
This study demonstrated the effectiveness of the proposed automatic breach-detection algorithm, which is based on real-time electrical conductivity measurements. Continuous monitoring across 51 porcine vertebrae enabled the system to stop drilling before spinal canal perforation, achieving a 100\% detection rate, with all cases graded \textit{"A"} or \textit{"B"} according to the Gertzbein-Robbins classification.

Electrical conductivity offers an X-ray-free approach to breach prevention that can complement navigation systems, reduce intraoperative imaging, and limit radiation exposure. In the current setup, the trajectory is defined by hand-guiding the robot. Integrating navigation would improve trajectory consistency and enable more systematic evaluation of breach detection under controlled conditions.

The proposed online, real-time breach detection method based on electrical conductivity could be integrated into standalone surgical tools or implemented as an additional safety feature in existing drilling systems. Further research may enable more advanced detection strategies and support deployment across different levels of autonomy, ranging from assisted manual drilling to semi-autonomous procedures, as demonstrated here, and ultimately to fully autonomous drilling systems designed for safe and precise execution.
%
%


\bibliography{sn-bibliography}

\end{document}